\documentclass[letterpaper, 10 pt, conference]{ieeeconf}
\IEEEoverridecommandlockouts
\usepackage{amsmath}
\usepackage{cite}
\usepackage{bbold}
\usepackage{epsfig}
\usepackage{hyperref}
\usepackage[caption=false,listofformat=subsimple,labelformat=simple]{subfig}

\usepackage{multirow}
\usepackage{multicol}
\usepackage{cellspace}
\newcolumntype{P}[1]{>{\centering\arraybackslash}p{#1}}

\title{\LARGE \bf
No More Potentially Dynamic Objects: Static Point Cloud Map Generation based on 3D Object Detection and Ground Projection}

\author{Soojin Woo*, Donghwi Jung*, and Seong-Woo Kim%
\thanks{* indicates equal contribution.\newline \indent All authors are with Seoul National University, Seoul, South Korea.\newline
 {\tt\footnotesize \{soojin.woo,donghwijung,snwoo\}@snu.ac.kr}}}
\begin{document}
\maketitle
\thispagestyle{empty}
\pagestyle{empty}

\begin{abstract}
In this paper, we propose an algorithm to generate a static point cloud map based on LiDAR point cloud data. Our proposed pipeline detects dynamic objects using 3D object detectors and projects points of dynamic objects onto the ground. Typically, point cloud data acquired in real-time serves as a snapshot of the surrounding areas containing both static objects and dynamic objects. The static objects include buildings and trees, otherwise, the dynamic objects contain objects such as parked cars that change their position over time. Removing dynamic objects from the point cloud map is crucial as they can degrade the quality and localization accuracy of the map. To address this issue, in this paper, we propose an algorithm that creates a map only consisting of static objects. We apply a 3D object detection algorithm to the point cloud data which are obtained from LiDAR to implement our pipeline. We then stack the points to create the map after performing ground segmentation and projection. As a result, not only we can eliminate currently dynamic objects at the time of map generation but also potentially dynamic objects such as parked vehicles. We validate the performance of our method using two kinds of datasets collected on real roads: KITTI and our dataset. The result demonstrates the capability of our proposal to create an accurate static map excluding dynamic objects from input point clouds. Also, we verified the improved performance of localization using a generated map based on our method. Our code is available at \url{https://github.com/woo-soojin/no_more_potentially_dynamic_objects}.
\end{abstract}

\section{Introduction}
As the use of 3D-related technologies spreads, research related to detecting 3D space with sensors and representing it as 3D data is actively underway. There are various methods for representing space in 3D, however, among them, point cloud is a representation method that depicts space with countless points consisting of three values: $x$, $y$, and $z$. Therefore, as the number of points increases, the information about space becomes richer, and it becomes possible to grasp more advanced spatial information, such as calculating normal vectors through clustering with surrounding points, in addition to basic coordinate information. Furthermore, as mentioned in references \cite{wang2018lidar,ladicky2017point,zhang2014loam,rozenberszki2020lol,peng2015obstacle}, the point cloud is essential in various fields including autonomous driving and robotics applications. Specifically, A point cloud can be utilized for the representation of 3D objects, the creation of meshes, a 3D scene reconstruction\cite{wang2018lidar}, and the digital twins that replicate real-world spaces in virtual environments.\cite{ladicky2017point} In autonomous driving and robotics fields, point cloud data is also applied for various applications such as map construction based on stacked point clouds \cite{zhang2014loam}, localization that supports vehicle and robot to find their current position \cite{rozenberszki2020lol}, and the path planning \cite{peng2015obstacle}. For the path planning task, the point cloud can be adopted to decide safe routes avoiding obstacles by applying an object detection algorithm that identifies surrounding objects.\\
\begin{figure}[t]
    \vspace{0.2cm}
    \centering
    \framebox{\parbox{0.49\textwidth}{\includegraphics[width=0.485\textwidth]{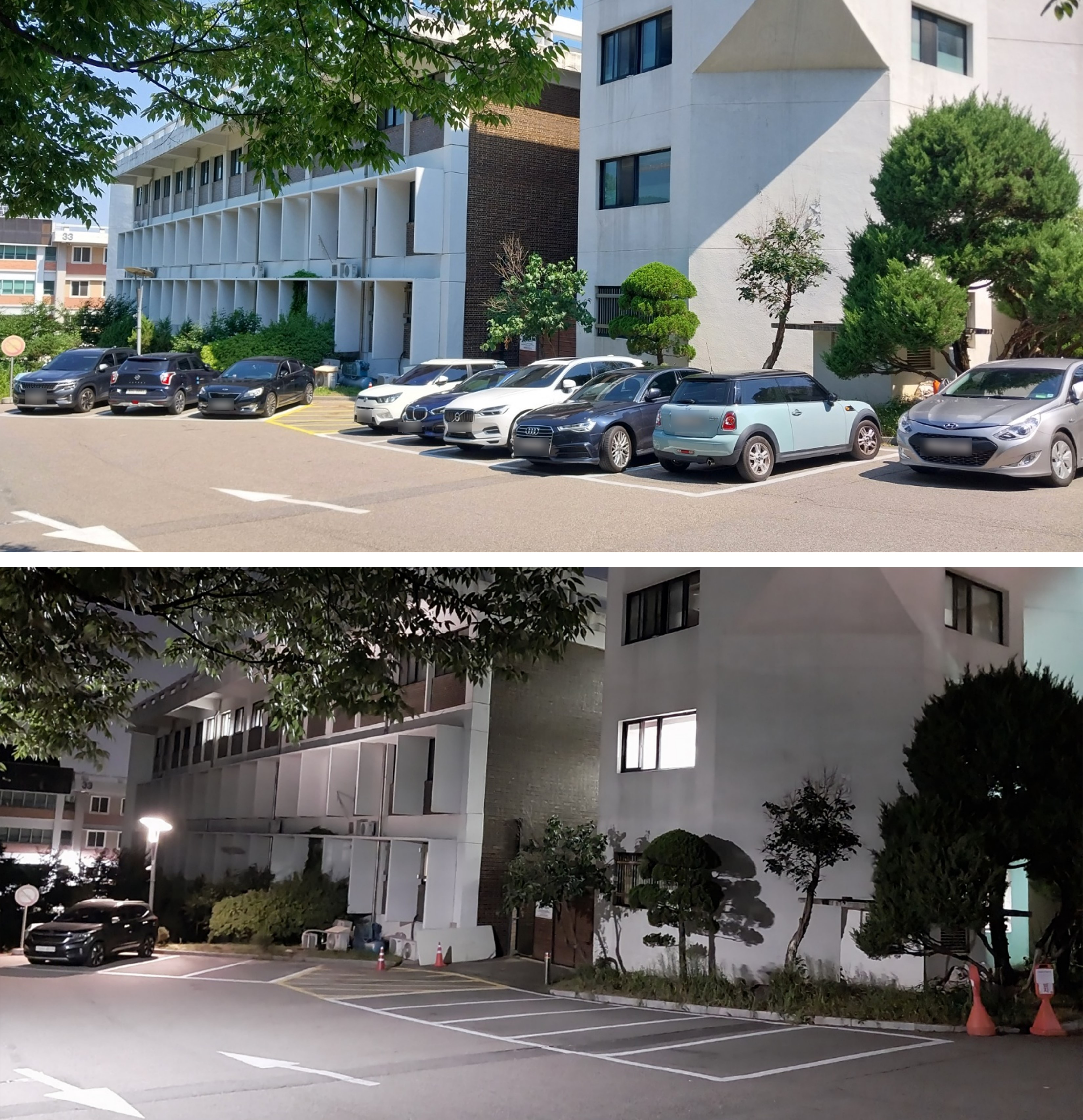}}}
    \caption{Changed distribution of potentially dynamic objects over time at the same location.}
    \label{fig:before_and_after_removal}
    \vspace{-0.4cm}
\end{figure}
\begin{figure*}[t]
    \vspace{0.2cm}
    \centering
    \framebox{\parbox{\textwidth}{\includegraphics[width=\textwidth]{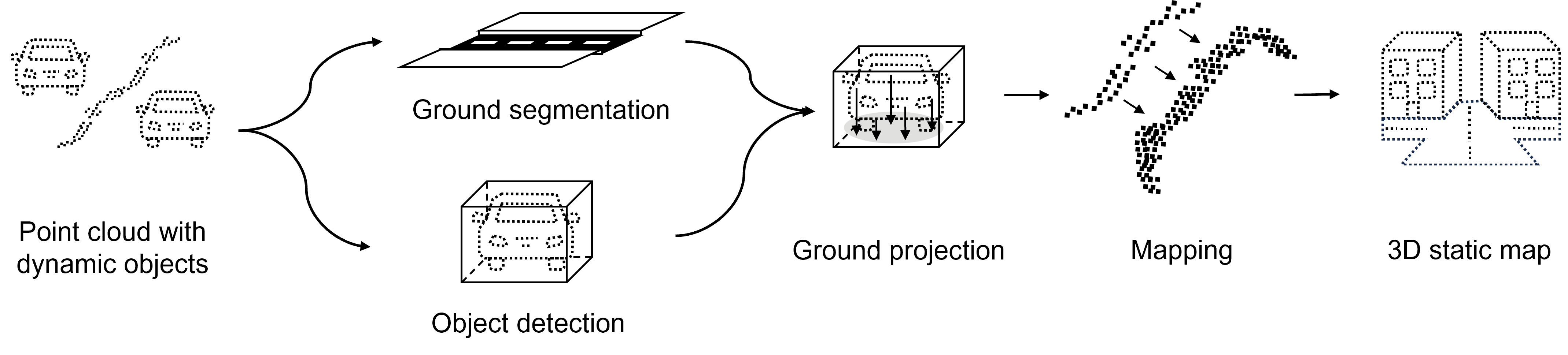}}}
    \caption{The process of our static map creation method.}
    \label{fig:process}
    \vspace{-0.4cm}
\end{figure*}
\indent In addition, a point cloud map is also one of the common utilization of point clouds. A point cloud map is generated by the accumulation of point clouds from sensors such as LiDAR using the estimated pose based on the algorithm \cite{zhang2014loam}. This point cloud map represents the 3D space based on the distributions of points containing information such as position and intensity. Among many kinds of point cloud maps, the point cloud maps used for autonomous driving are mostly generated from the data collected on the roads. Therefore, the data is filled with dynamic objects such as vehicles and pedestrians. And the distribution of these objects varies over time as shown in Fig. \ref{fig:before_and_after_removal} by changing their locations. The movement of dynamic objects not only degrades the accuracy of scan-to-map matching in localization but also can lead to inefficient path planning by mistakenly perceiving the areas as impassable even though the place is actually empty. For these reasons, the representation of surroundings only with static objects having fixed positions is important for the accuracy of algorithms including localization and path planning. In other words, dynamic objects in the 3D point cloud map must be identified and eliminated.\\
\indent Removing dynamic objects, as mentioned above, is a crucial issue in autonomous driving and robotics. However, at the moment of map creation, it is difficult to distinguish static objects and dynamic objects. In previous research that removes dynamic objects from the map includes studies such as \cite{hornung2013octomap,schauer2018peopleremover,pagad2020robust,kim2020remove,fan2022dynamicfilter,lim2021erasor,milioto2019rangenet++,chen2021moving,chen2022exploring}. These studies differentiate dynamic objects from static objects, and remove them predicting the dynamic objects that moved at the time of data acquisition. Therefore, this approach can only identify and eliminate objects that changed their positions at the time of data collection. However, there is a limitation in that potentially dynamic objects, such as parked cars, can not be recognized as dynamic objects. Consequently, they are not removed from the map because these objects are identified as static objects during the mapping process. As shown in Fig. \ref{fig:before_and_after_removal}, the potentially dynamic objects that can change their locations over time are indistinguishable from static objects such as buildings and trees during the data collecting process. However, if the dynamic objects move, they might not be visible when the data is acquired next time. Therefore, it is necessary to remove potentially dynamic objects from the map, however this is a challenging task to resolve with existing methods. This limitation not only decreases the quality of the static map but also the performance of algorithms including localization and path planning which rely on the static map.\\
\indent In this paper, we propose a pipeline that detects objects based on their features in a single frame rather than referring to its time-dependent features and then removes dynamic objects by projecting them to the calculated ground. Methods for detecting objects using their traits have been extensively studied in various fields, including camera-based approaches \cite{girshick2015fast, redmon2016you} and LiDAR-based approaches \cite{zhou2018voxelnet,simony2018complex}. When using data collected by the camera, objects are found in images, classified, and then 2D bounding boxes are obtained. Considering the sparsity of LiDAR data, the methods based on LiDAR data convert point cloud into various formats to improve the detection performance. The \cite{zhou2018voxelnet} converts point cloud data into voxels for 3D object detection. In \cite{simony2018complex}, the input point cloud data is converted to a 2D Birds-Eye-View (BEV) RGB-map to get the coordinates of the object's bounding box.\\
\indent Among these methods, we employed a voxel-based method for 3D object detection which adopts LiDAR point cloud as input data. Using this approach, we detect surrounding 3D objects and distinguish dynamic and static objects based on returned classes on each object. The objects classified as dynamic are projected to the ground plane calculated by the ground segmentation algorithm. The datasets used in this paper were collected on the road which means, the majority of dynamic objects such as cars exist over the road. Considering this trait, we conducted ground projection on dynamic objects to replace the car with the road as if there was no car on the road in the first place. Subsequently, we create a static point cloud map with point cloud data without dynamic objects. As a consequence, the static map constructed by our method only includes static objects by removing both dynamic and potentially dynamic objects from the map which makes our approach independent from data acquisition timing.\\
\indent To the best of our knowledge, our research is the very first to propose a process that utilizes 3D object detection and ground projection algorithms on a point cloud to generate a static point cloud map by removing potentially dynamic objects. We confirmed the superior localization performance when we used static maps created by our methods compared to maps generated by existing approaches.\\
\indent The contributions of our paper are as follows:
\begin{itemize}
\item \emph{A Static Map Creation Process}: We proposed a process that connects 3D object detection and ground projection to remove potentially dynamic objects and generate a static point cloud map.
\item \emph{Public Release of Code}: We made the code used in our research publicly available, enabling other researchers to utilize it.
\end{itemize}
\begin{figure*}[t]
\vspace{0.2cm}
    \centering
    \framebox{\parbox{\textwidth}{\includegraphics[width=\textwidth]{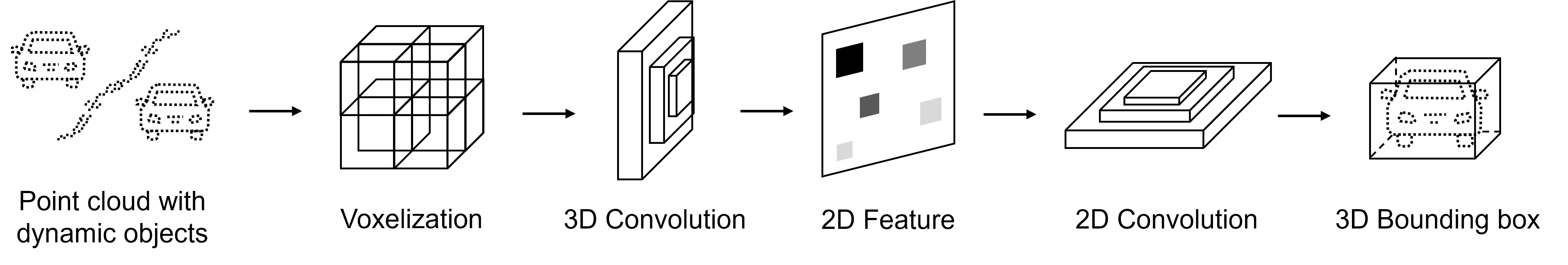}}}
    \caption{The process of voxel-based 3D object detection.}
    \label{fig:network_architecture}
\end{figure*}

\section{Related Works}
There are methods for detecting and removing dynamic objects that exploit occupancy maps. This approach uses ray tracing to evaluate the occupancy of a specific area and determines the presence of dynamic objects by inspecting the continuity of occupancy between consecutive frames. While this approach demonstrates high accuracy in dynamic object removal, it imposes a significant computational burden, particularly in occupancy calculations using ray tracing. To address this computational burden, previous research, such as Hornung \emph{et al.}'s work \cite{hornung2013octomap}, attempted to enhance optimization. However, as seen in the paper \cite{schauer2018peopleremover}, computational challenges remained. Additionally, Pagad \emph{et al.} \cite{pagad2020robust} improved performance by applying object detection to Octomap \cite{hornung2013octomap}, however computational burdens still persisted.\\
\indent To mitigate the computational challenges posed by occupancy map-based methods, visibility-based methods were proposed. Unlike occupancy map-based methods, this approach focuses on the visibility of objects, reducing the computational load. For instance, when a specific object is occluded by another object, it is considered a dynamic point, reducing the computational requirements compared to occupancy map-based methods. However, this simplification leads to decreased accuracy, resulting in frequent false positives. To address this issue, Kim \emph{et al.} \cite{kim2020remove} introduced an iterative refinement process to remove false positives. Nevertheless, challenges, such as less accuracy, still exist in visibility-based methods. Therefore, Fan \emph{et al.} \cite{fan2022dynamicfilter} proposed a method that integrates both visibility-based and map-based approaches to achieve high accuracy and less computational burden in dynamic object removal. However, their method is limited to detecting currently dynamic objects. In other words, Dynamicfilter \cite{fan2022dynamicfilter} could not detect the potentially dynamic objects such as parked cars.\\
\indent There are also model-based methods for removing dynamic objects using a model. Among these, there is research that utilizes ground models including the paper \cite{lim2021erasor}. In this paper, the authors removed dynamic objects by assuming that all dynamic objects are in contact with the ground. However, this research also relies on the motion characteristics of dynamic objects to identify them, potentially dynamic objects that may not change their positions over a short period. Additionally, previous methods \cite{milioto2019rangenet++,chen2021moving} that adopted deep learning-based approaches for dynamic object removal identified objects based on features such as shape rather than their motion. Therefore, these method could detect dynamic objects including potentially dynamic objects. However, these methods projected the point cloud into 2D range images and identified dynamic objects in these images. This projection selected one point per pixel, resulting in information loss and reduced object detection performance. To address this issue, in our paper, instead of projecting the point cloud into 2D range images, we use the shape of 3D voxels to ensure that more points are kept. Moreover, Chen \emph{et al.} \cite{chen2022exploring} proposed a LiDAR SLAM algorithm using deep learning-based dynamic object removal. In their study, they utilized the PointPillars \cite{lang2019pointpillars} to find dynamic objects in 3D point clouds without 2D transformations. However, they removed points corresponding to dynamic objects, which reduced accuracy of LiDAR SLAM due to sparsity of point clouds. Therefore, in our paper, we improve mapping performance by utilizing ground projection through ground segmentation, preserving points instead of removing them.\\
\indent In this paper, we develop a model-based approach capable of detecting both currently dynamic objects and potentially dynamic objects. To achieve higher accuracy in obtaining the 3D bounding box, we opt for a voxel-based object detection method in our process. Subsequently, after projecting the points associated with dynamic objects onto the ground, we generate a highly accurate static map through mapping.

\section{Methods}
In this section, we will describe the process, as depicted in Fig. \ref{fig:process}, which removes dynamic components from a given 3D point cloud map to create a static point cloud map. First, we transform the point cloud data from the LiDAR sensor into voxel features. We perform dynamic object detection by providing transformed data as input to a CNN-based neural network. Additionally, we apply a ground segmentation algorithm to project data to the ground after calculating the ground plane based on raw point cloud data. Specifically, we use 3D bounding boxes corresponding to dynamic objects returned by a 3D object detector to find out the points inside the boxes to project them on the ground. Finally, we stack these projected point clouds through mapping to create the 3D static point cloud map. The following steps will provide sequential explanations of our process.
\subsection{3D Object Detection}\label{method:object_detection}
There are several methods for 3D object detection, including the use of mathematical equations. Among them, in this paper, we utilize the neural network-based method that has been widely studied recently. The reason for using the neural networks is that we can leverage GPUs to process incoming point clouds in real-time with high accuracy. However, unlike data such as camera image data, the point clouds coming from LiDAR do not have a consistent size for each data sequence. Therefore, applying data embedding using neural networks, such as convolution layers, is challenging. To overcome this limitation, in this paper, we adopt a 3D object detection method that utilizes spaced voxels based on subdivided 3D space. Voxels are data with fixed dimensions in the shape of 3D cubes, allowing for embedding using 3D convolution. As a result of this embedding, 2D features are obtained, and applying 2D convolution layers ultimately allows us to predict 3D bounding boxes. At this point, we set the prediction values through the network as the center position, size, and orientation angle of the bounding box, ultimately obtaining the 3D bounding box. Then, we use this obtained bounding box to filter the point cloud. As a result, we can identify the points corresponding to dynamic objects. The process of 3D object detection is described in Fig. \ref{fig:network_architecture} and formulated as follows:
\begin{gather}
   p=(x,y,z),\\
   P=\{p\},\\
   \bar{P}=\{\bar{p}|\theta^{min}\leq\bar{p}\leq\theta^{max}\},\\
   \bar{p}\in V_{ijk}\;(\text{if}\;\tau^{min}_{ijk}\leq\bar{p}\leq\tau^{max}_{ijk}),\\
   (x_v,y_v,z_v)=\left(x_c\pm\frac{w}{2},y_c\pm\frac{l}{2},z_c\pm\frac{h}{2}\right),\\
    R=\begin{pmatrix}
    \cos{\theta}&-\sin{\theta}&0\\
    \sin{\theta} & \cos{\theta}& 0\\
    0&0&1
    \end{pmatrix},\\
    (\bar{x_e},\bar{y_e},\bar{z_e})^\top=R\cdot(x_v,y_v,z_v)^\top,
\end{gather}
where $P$ represents the incoming point clouds, and $p$ represents the individual points composing the point clouds, with coordinates of $x$, $y$, and $z$. In addition, $\theta^{min}$ and $\theta^{max}$ indicate the minimum and maximum thresholds for filtering the point clouds in $x$, $y$, and $z$ for voxelization. The points that are included in the voxel being passed through the filtering represent $\bar{p}$, while $\bar{P}$ signifies the point clouds composed of these $\bar{p}$ points. Additionally, $V_{ijk}$ corresponds to the $i$th, $j$th, and $k$th voxel in the $x$, $y$, and $z$ directions, respectively. $\tau^{min}_{ijk}$ and $\tau^{max}_{ijk}$ represent the thresholds for the $x$, $y$, and $z$ values that determine the points included in that specific voxel. Furthermore, $x_c$, $y_c$, and $z_c$ denote the $x$, $y$, and $z$ values which are the center position of the bounding box. Moreover, $w$, $l$, and $h$ indicate the width, length, and height of the 3D bounding box. On the $x-y$ plane, $\theta$ is the orientation of the bounding box. $x_c$, $y_c$, $z_c$, $w$, $l$ and $h$ are obtained as an output of 3D object detection model. On top of that, $x_v$, $y_v$, and $z_v$ mean the $x$, $y$, and $z$ values expressing the positions of the bounding box's vertices. Additionally, $R$ indicates the rotation matrix based on $\theta$. Lastly, $\bar{x}_e$, $\bar{y}_e$ and $\bar{z}_e$ are the transformed positions of bounding box's vertices.
\subsection{Ground Segmentation}\label{method:ground_extration}
Simply removing the points of dynamic objects consequently reduces the number of points that are matched to the previously accumulated scans during the scan-to-map matching stage, and it leads to the decreased accuracy of LiDAR mapping by diminishing the number of points that can be useful for LiDAR mapping. Therefore, we maintained the total number of points by projecting the points corresponding to dynamic objects onto the ground plane in the middle of our dynamic object removal process to sustain the performance of the LiDAR mapping. To execute the projection, it is crucial to determine the equation on the target plane, which in this case is the ground plane. Assuming that the vehicle is on the ground and the slope is minimal during motion, the ground plane will be parallel to the $x$-$y$ plane in the world coordinate system. As a result, the normal vector of the ground plane will have values only in the $z$ direction. To project dynamic object points onto the ground plane, the average $z$ value of the ground plane denoted as $\bar{z}$ needs to be calculated. In our paper, we achieve the ground plane calculation by performing ground segmentation targeting the entire input point clouds, isolating the points identified as ground, and then calculating the average $z$ value of these points. This can be mathematically expressed as follows:
\begin{gather}
    P_{gr}=\{p_i|i \in I_{gr}\},\\
    P=P_{ng} \cup P_{gr},\\
    \bar{z}=\frac{1}{n}\cdot\Sigma^n_{k=1} P_{gr},
\end{gather}
where $P_{gr}$ represents the set of points in the point clouds that belong to the ground, and $I_{gr}$ is the set of indices of points in $P$ which are part of $P_{gr}$. On the other hand, $P_{ng}$ denotes the set of points in the point clouds which are not included in the ground. Additionally, $n$ expresses the number of ground points.
\subsection{Projection and Mapping}
\begin{figure}[t]
    \vspace{0.15cm}
    \centering
    \framebox{\parbox{0.49\textwidth}{\includegraphics[width=0.49\textwidth]{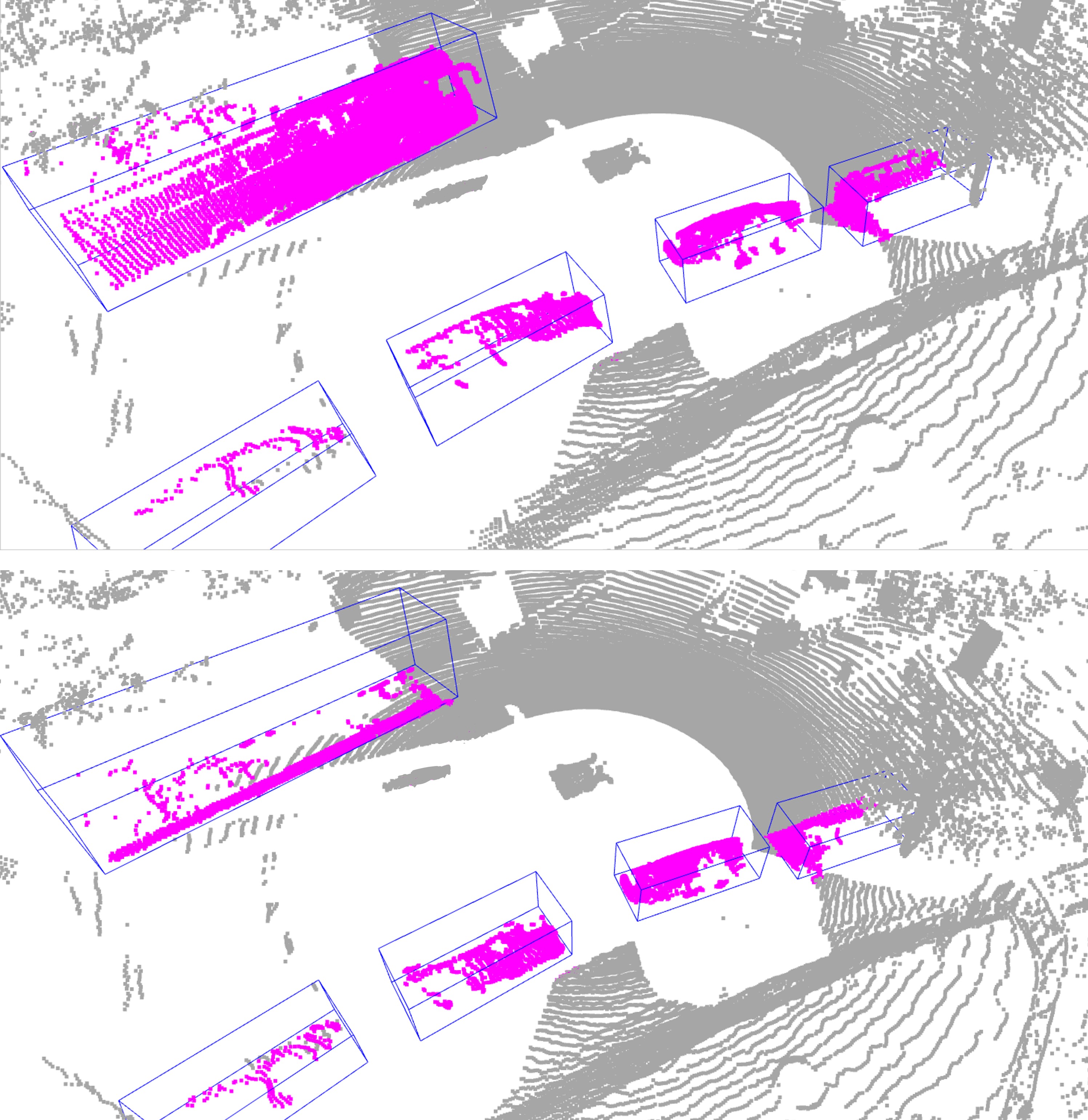}}}
    \caption{The projection result of the point cloud onto the ground plane within the 3D bounding box obtained by the 3D object detection model. Dynamic objects (pink), Static objects (gray).}
    \label{fig:projection_before_and_after}
\end{figure}

We utilize the bounding boxes obtained from 3D object detection process in Sec. \ref{method:object_detection} to find out the point cloud and extract points corresponding to dynamic objects. Subsequently, we use the mean $z$ value of the ground, $\bar{z}$, calculated in Sec. \ref{method:ground_extration}, to project the labeled points as dynamic objects onto the ground. The appearance before and after projection is shown in Fig. \ref{fig:projection_before_and_after}. The points corresponding to the dynamic objects are projected onto the ground through ground segmentation, and then we create the final 3D point cloud map using 3D LiDAR Simultaneous Localization and Mapping (SLAM). Specifically, the data which is composed of projected points, and the map which is accumulated until the previous step are used as input data of SLAM. During this process, loop closure is used to correct prediction errors resulting from long-distance driving. This procedure can be expressed mathematically as follows:
\begin{gather}
    \hat{p_i}=
    \begin{cases}
        (x_i,y_i,z_i), & \text{if }i \in I_{gr},\\
        (x_i,y_i,\bar{z}), & \text{otherwise},
    \end{cases},\\
    \hat{P}=\{\hat{p}\},\\
    \hat{P}_{k+1} \oplus M_k = M_{k+1},
\end{gather}
where $\hat{p}$ represents points that have been projected onto the ground plane, and $\hat{P}$ denotes the point cloud composed of $\hat{p}$. Additionally, $M$ represents the point cloud map. Furthermore, $\oplus$ signifies the mapping process utilizing LiDAR SLAM.
\section{Experiments}
\subsection{Implementation}
In this paper, we utilized VoxelNeXt \cite{chen2023voxelnext} for 3D object detection. For ground segmentation, we applied the Patchwork++ algorithm \cite{lee2022patchwork}. Point cloud mapping was performed using LOAM \cite{zhang2014loam}, while loop closure utilized Scan Context \cite{kim2018scan}. We assessed the performance of localization based on the accuracy of scan-to-map matching using Iterative Closest Point (ICP) registration.

\subsection{Data}
We evaluated the performance of the proposed algorithm that generates the static map both in a qualitative and quantitative way using datasets, such as KITTI \cite{geiger2012we} and ours. Additionally, our dataset was acquired following the same path on different dates to validate the performance of mapping and localization according to the existence of dynamic objects on the same path. For the evaluation process, we selected two types of datasets. One contained a number of dynamic objects, while the other one had comparatively fewer dynamic objects by applying a dynamic object removal algorithm. The reason for varying the distribution of dynamic objects is that our research aims to verify whether autonomous driving algorithms, such as mapping and localization, are affected by changing distributions of dynamic objects over time. In the mapping process, we utilized the dataset with abundant dynamic objects to prove the ability of our algorithm that remove dynamic objects from the map. For the localization task, we used the 
data that dynamic objects have been removed to verify whether the algorithm performs well in different environments, apart from the mapping stage.
\subsection{Evaluation}
We proved the capability of our algorithm in quantitative and qualitative ways. In Sec. \ref{qualitative_analysis}, we qualitatively compared the results of mapping using a map created by our algorithm and by other methods. Ground truth denotes the map created using GPS information, and without removal expresses the map constructed without applying a dynamic object removal algorithm.

For quantitative analysis, in Sec. \ref{quantitative_analysis}, we measured the accuracies of localization on each static map generated by dynamic object removal algorithms including our methods. As reference methods, we applied Removert\cite{kim2020remove} and Erasor\cite{lim2021erasor}.
Additionally, we established the baseline by not applying any dynamic object removal method for performance comparisons. For the localization performance evaluation, we measured the accuracy of localization at time $k+1$ by performing ICP registration between the 3D point cloud map and the point cloud at time $k+1$. We used the ground truth pose at time $k$ as the initial transformation for global registration. Then, we calculated the Root Mean Squared Error (RMSE) of the relative transformation between the ground truth poses at time $k$ and time $k+1$ for the entire path. The results are shown in Table \ref{table:localization_result}.

\begin{table}[t]
\vspace{0.2cm}
\small
\caption{Comparison of SLAM odometry accuracy.\\
(Average error; meter).}
\label{table:slam_result}
\begin{center}
    \renewcommand{\arraystretch}{2}
    \begin{tabular}{P{0.10\textwidth} P{0.05\textwidth} P{0.10\textwidth} P{0.10\textwidth}}\hline
    &seq&w/o removal&Our method\\\hline
    \multirow{4}{*}{KITTI\cite{geiger2012we}}&00&\textbf{1.60}&2.68\\
    &05&\textbf{2.40}&2.48\\
    &07& 1.55&\textbf{1.52}\\
    &08& 6.42&\textbf{5.97}\\\hline
    Our data&&\textbf{3.56}&4.33\\\hline
    \end{tabular}
\end{center}
\end{table}

\begin{table}[t]
\small
\caption{Comparison of localization performance.\\
(Average error; meter, radian).}
\label{table:localization_result}
\begin{center}
\renewcommand{\arraystretch}{2}
    \begin{tabular}{P{0.10\textwidth} P{0.06\textwidth} P{0.10\textwidth} P{0.10\textwidth}}\hline
    &&w/o removal&Our method\\\hline
    \multirow{2}{*}{Our data}&x-y&0.950&\textbf{0.900}\\
    &yaw&0.070&\textbf{0.066}\\\hline
    \end{tabular}
    \vspace{-0.4cm}
\end{center}
\end{table}

\subsubsection{Mapping}\label{qualitative_analysis}
\begin{figure}[t]
    \vspace{0.15cm}
    \centering
    \framebox{\parbox{0.49\textwidth}{\includegraphics[width=0.49\textwidth]{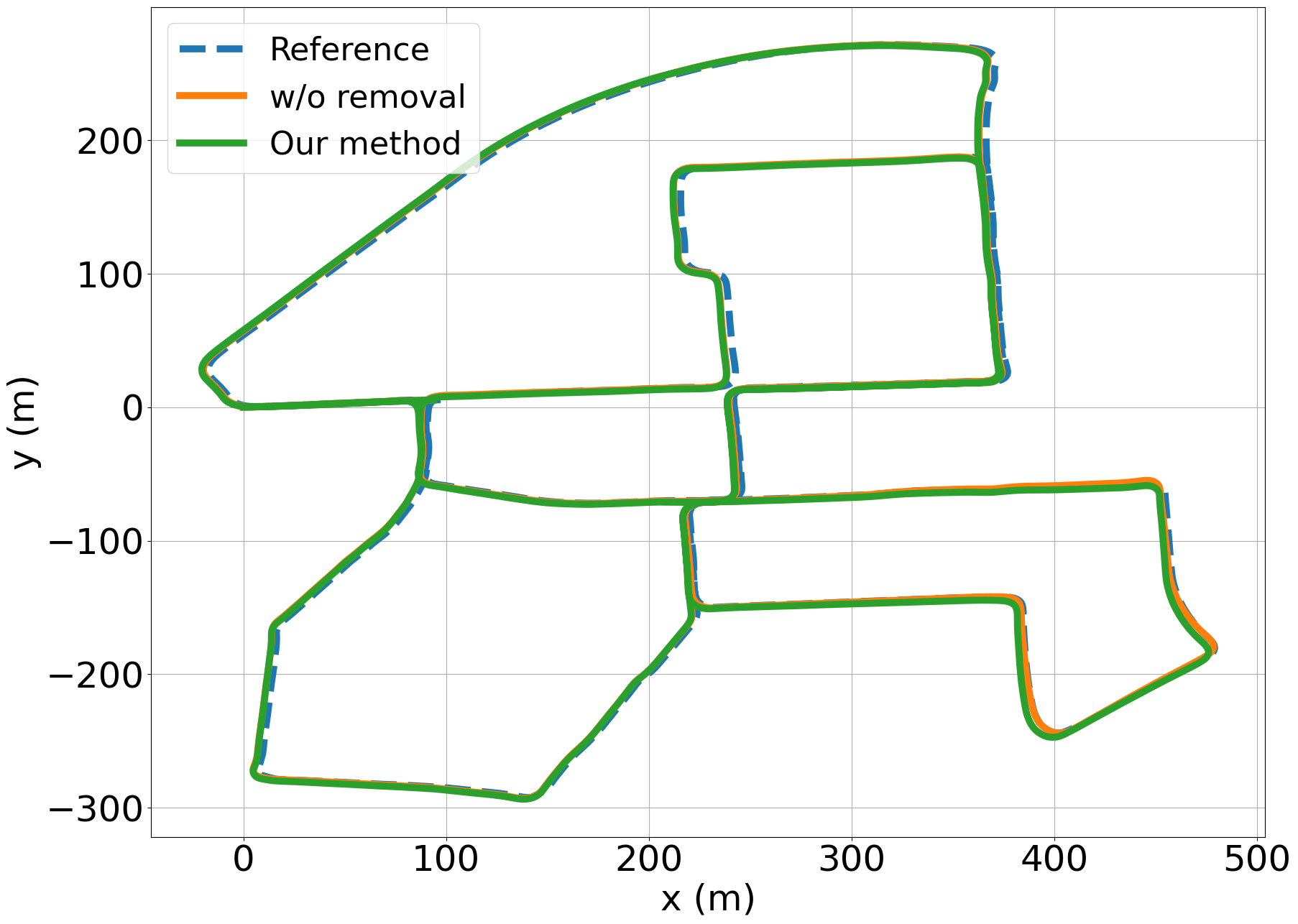}}}
    \caption{Comparison of SLAM odometry prediction performance on the 00 sequence of KITTI dataset.}
    \label{fig:kitti_result}
\end{figure}

\begin{figure}[t]
    \centering
    \framebox{\parbox{0.49\textwidth}{\includegraphics [width=0.49\textwidth]{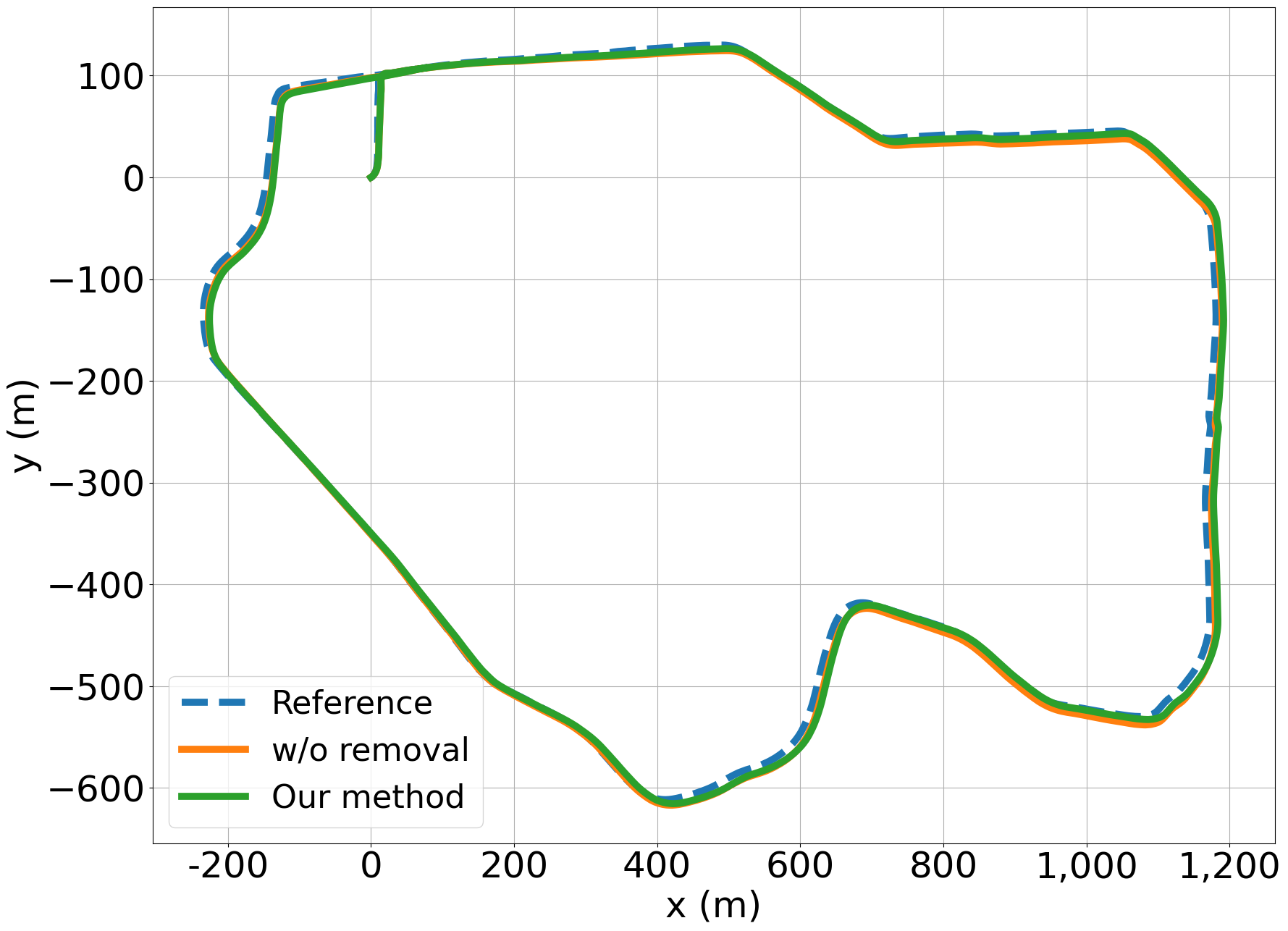}}}
    \caption{Comparison of SLAM odometry prediction performance on our dataset.}
    \label{fig:snu_result}
\end{figure}

Fig. \ref{fig:kitti_result} and \ref{fig:snu_result} show the results that applied our algorithm on KITTI and the dataset of our campus. Firstly, we detected dynamic objects and then projected points composing dynamic objects to the ground. Finally, we adapted LiDAR SLAM for the mapping process to validate the capability of our proposal. As shown in Table \ref{table:slam_result}, we compared the mapping results for each dataset with the results of mapping using the raw point cloud before ground projection. By comparing the mapping results by our method with the raw point cloud-based result, we demonstrated that the performance of mapping is maintained even after ground projection. As demonstrated in Fig. \ref{fig:kitti_result}, \ref{fig:snu_result} and Table \ref{table:slam_result}, we showed high accuracy of mapping results for all two datasets including KITTI and our campus dataset. Therefore, we confirmed that ground projection for dynamic objects maintains mapping performance.
\begin{figure}[t]
    \vspace{0.15cm}
    \centering
    \framebox{\parbox{0.49\textwidth}{\includegraphics[width=0.485\textwidth]{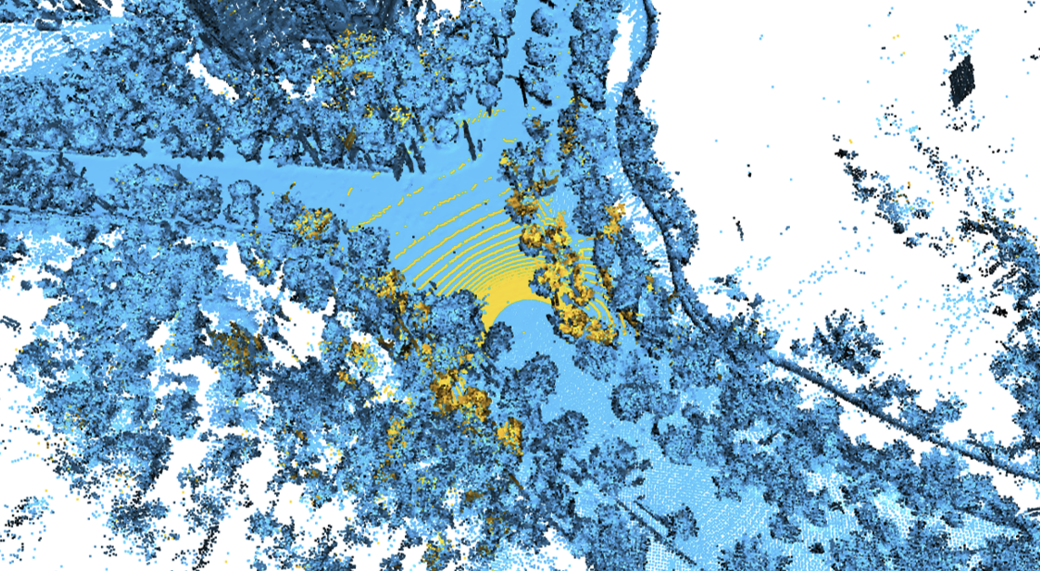}}}
    \caption{Scan to map matching. Scan (yellow), Map (Cyan).}
    \label{fig:scan_map_matching}
\end{figure}

\subsubsection{Localization}\label{quantitative_analysis}
For localization, we calculated the performance using the accuracy of ICP-based registration between the scan and map, as shown in Fig. \ref{fig:scan_map_matching}. Table \ref{table:localization_result} shows the localization error with the unit of meter. We demonstrated the superior performance of our algorithm by showing lower localization errors compared to the baseline (without removal). We achieved higher localization accuracy by using our proposal to remove potentially dynamic objects from the data used as input for the localization.

\section{Conclusion}
In our paper, we proposed a novel and robust method for removing dynamic objects from a given point cloud map through 3D object detection and ground projection. The previous research relied on temporal information, therefore they could only eliminate currently moving objects from their map. However, our method not only removed moving objects at the time of data acquisition but also potentially dynamic objects that change their location across time. Rather than simply removing the points from the map that compose dynamic objects, we projected corresponding points to the ground plane during the mapping stage to maintain the performance of mapping and localization. In particular, we found the dynamic objects based on the results of the 3D object detection algorithm. To calculate the average height value of the ground plane, we applied ground segmentation to the raw point clouds. With an estimated value by the ground segmentation process, we projected the points corresponding to the dynamic objects based on the results of the detection algorithm. Subsequently, we used previously gained point cloud data by projecting dynamic points as input of the LiDAR mapping process to create a static map, dynamic objects are completely erased. The static map constructed by our method does not possess both currently moving and potentially dynamic objects. We evaluated the mapping performance of our proposal with two types of datasets. One is publicly available and the other one was acquired by ourselves. We also verified the localization performance using the generated map with our method. Through evaluation, we confirmed the robustness of our pipeline.

\section*{Acknowledgement}
This work is supported by the National Research Foundation of Korea (NRF) through the Ministry of Science and ICT under Grant  2021R1A2C1093957, by the Korean Ministry of Land, Infrastructure and Transport(MOLIT) as "Innovative Talent Education Program for Smart City", and by Korea Institute for Advancement of Technology(KIAT) grant funded by the Korea Government(MOTIE) (P0020536, HRD Program for Industrial Innovation). The Institute of Engineering Research at Seoul National University provided research facilities for this work.
\bibliographystyle{IEEEtran}
\bibliography{root}
\end{document}